\documentclass[10pt,twocolumn,letterpaper]{article}

\usepackage{iccv}
\usepackage{times}
\usepackage{epsfig}
\usepackage{graphicx}
\usepackage{amsmath}
\usepackage{amssymb}
\usepackage{caption}
\usepackage{subcaption}
\usepackage{authblk}

\usepackage[pagebackref=true,breaklinks=true,letterpaper=true,colorlinks,bookmarks=false]{}

\def\beq{\begin{equation}}
\def\eeq{\end{equation}}
\def\norm{||}

\iccvfinalcopy 

\def\iccvPaperID{862} 

\ificcvfinal\fi
\begin{document}

\title{Pose Embeddings: A Deep Architecture for Learning to Match Human Poses}

\author[1]{Greg Mori\thanks{This work was done while Greg Mori was a visiting scientist at Google Inc.}}
\author[2]{Caroline Pantofaru}
\author[2]{Nisarg Kothari}
\author[2]{Thomas Leung}
\author[2]{George Toderici}
\author[2]{Alexander Toshev}
\author[2]{Weilong Yang}
\affil[1]{Simon Fraser University}
\affil[2]{Google Inc.}


\maketitle

\begin{abstract}
We present a method for learning an embedding that places images of humans in similar poses nearby.  This embedding can be used as a direct method of comparing images based on human pose, avoiding potential challenges of estimating body joint positions.  Pose embedding learning is formulated under a triplet-based distance criterion.  A deep architecture is used to allow learning of a representation capable of making distinctions between different poses.  Experiments on human pose matching and retrieval from video data demonstrate the potential of the method.
\end{abstract}

\section{Introduction}
\begin{figure}
\includegraphics[trim=0cm 10cm 12cm 4cm, clip=true, width=0.5\textwidth]{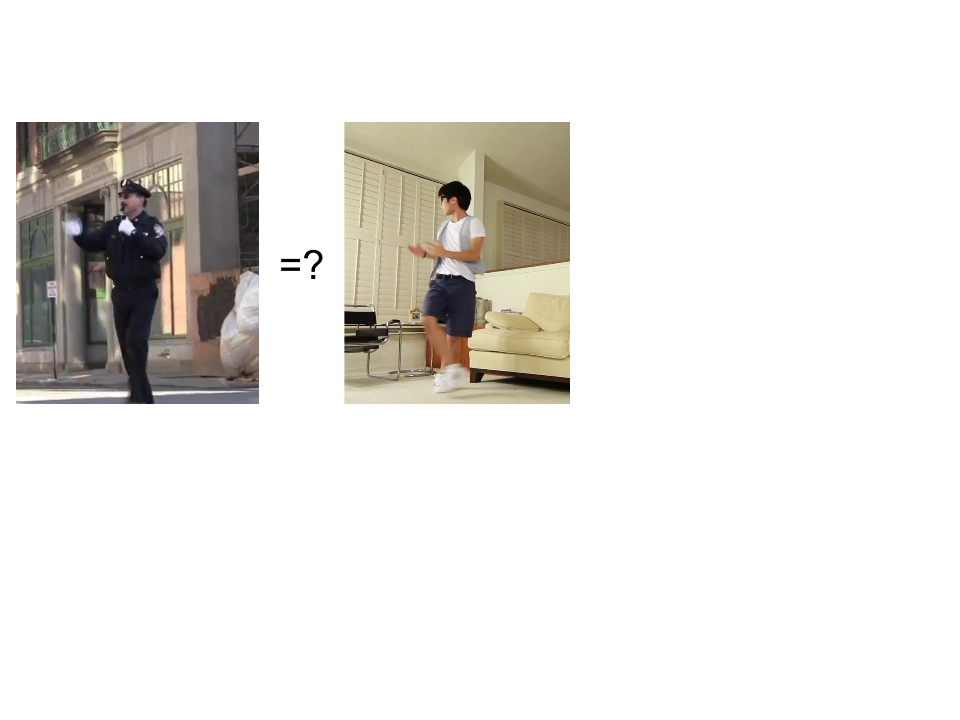}
\includegraphics[trim=0cm 10cm 0cm 4cm, clip=true, width=0.5\textwidth]{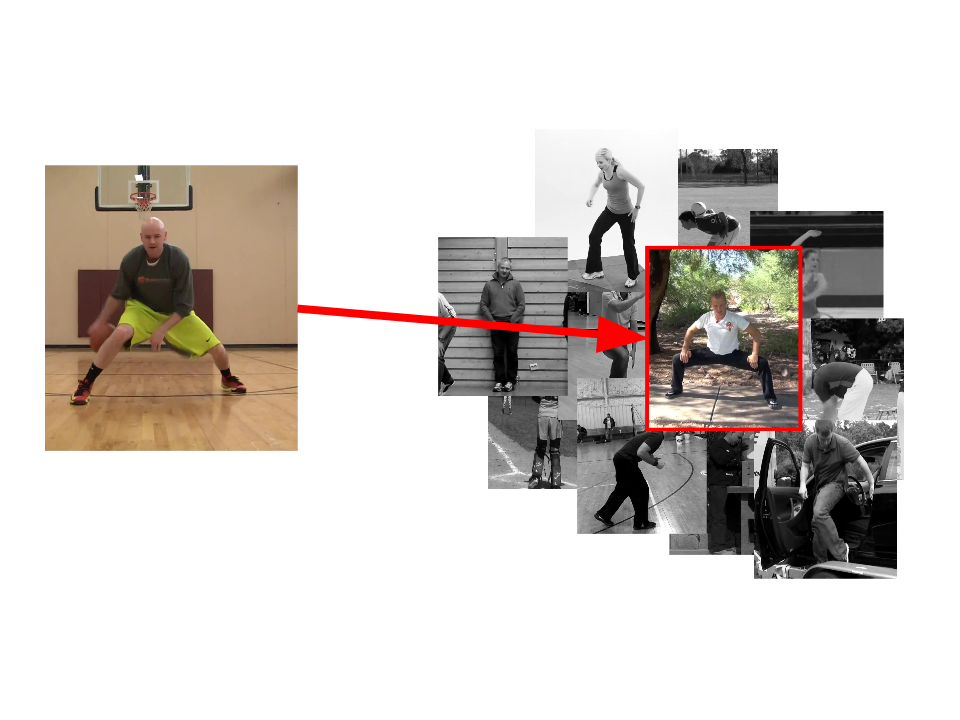}
\caption{We learn a human pose embedding space that places images of people similar poses nearby.  Top: pose embeddings can be used to compare two images based on pose. Bottom: retrieval of similar pose images from a database.}
\label{fig:intro}
\end{figure}

Are two people in similar poses?  Consider the image examples in
Fig.~\ref{fig:intro}.  Answering this question can be done in
different ways.  A standard approach is to perform human pose
estimation, localizing the positions of a set of body joints.  Given
these body joint positions, a similarity measure over poses could be
defined.

As an alternative, in this paper we develop a direct method for
comparing human poses, obviating the need for explicit pose
estimation.  We learn an embedding that aims to place images of people
in similar poses near each other.

This direct embedding method possesses several advantages.  First, it
avoids the challenging problem of localizing individual joints.  In
spite of great progress in human pose estimation methods, occluded
parts and unusual poses remain confounding factors.  Further, pose
estimation methods require choosing a representation for human pose,
such as a fixed set of body part locations.  The sufficiency of this
representation for representing pose similarity and its utility in
situations of occlusion are problematic.  Finally, directly learning
pose similarity permits modeling regions of pose space and learning
sensitivity to pose differences of varying magnitude over this space.

Providing an automated algorithm to answer the pose similarity
question enables a variety of applications. Pose search can be used in
a query-by-example video retrieval setting for example: given an image
of a person in a pose, find similar posed people in frames of a video
collection.  Group activity analysis, labeling the sets of people who
are interacting in a scene, can be done based upon similarity in pose:
people engaged in conversation or having a meal together tend to share
commonalities in pose.

The contribution of this paper is the formulation of human pose
matching as a direct learning problem based on a deep architecture. We
present an algorithm for learning pose matching from simple pose
similarities, potentially avoiding the need for detailed labeling of
human poses when learning pose retrieval.  We demonstrate the
generality of these learned pose representations by transfering the
learned models to pose-based video retrieval and group activity
clustering.

\section{Previous Work}
In this paper we develop a method for learning human pose similarity.

{\bf Human pose exemplar matching}: Template matching approaches have
deep roots in the computer vision literature.  Early work on
exemplar-based matching methods for human detection and pose
estimation relied on edge or silhouette
detection. Gavrila~\cite{gavrila2000pedestrian} performed Chamfer
matching of edge maps and organized human poses in a hierarchy.  Mori
and Malik~\cite{MoriM02} matched using shape descriptors.
Shakhnarovich et al.~\cite{ShakhnarovichVD03} generated large volumes
of synthetic exemplar images.  Lin et al.~\cite{lin2009recognizing}
developed hierarchies of exemplars, represented with both shape and
motion features.

{\bf Model-based pose estimation}: The pictorial structure
model~\cite{felzenszwalb2005pictorial} has formed the backbone for a
number of successful methods for pose estimation.  Ferrari et
al.~\cite{FerrariMZ09} explored the use of pictorial structure pose
estimation models for pose search.  Yang and Ramanan~\cite{YangR11}
extended this model to large number of small, flexibly arranged
parts. Johnson and Everingham focused on challenging poses via mixture
models and cleaning up training data annotations~
\cite{JohnsonE11}. Ionescu et al.~\cite{IonescuLS11} use image
segmentation and model valid regions of pose space around examples.
Sapp and Taskar~\cite{SappT13} develop efficient approaches for
utilizing non-tree strutured models Pishchulin et
al.~\cite{PishchulinAGS13} also expand modeling ability, conditioning
models on poselets~\cite{BourdevM09}, clustered body part examples.

Recent state of the art methods have used deep learning for pose
estimation.  Toshev and Szegedy~\cite{ToshevS14} formulated pose
estimation as a regression problem and included a coarse-to-fine
strategy to refine estimates from the deep network.  Tompson et
al.~\cite{TompsonJLB14} estimated individual body joint locations
which are then combined in a message passing-style network.

{\bf Pose spaces}: Previous work on pose spaces includes learning
manifolds for human pose: methods that regressed human pose from input
images.  Urtasun et al.~\cite{urtasun20063d} pioneered the use of
latent variable spaces, for example using them to track human figures
in video sequences.  Pavlovic~\cite{Pavlovic04} also focused on motion
sequences, learning densities over human motions with applications to
motion clustering.  Agarwal and Triggs~\cite{agarwal20043d} formed
direct regression from silhouette features to human pose.  Athitsos et
al.~\cite{AthitsosASK04} form embedding spaces that permit efficient
retrieval, demonstrating the ability to retrieve hand poses and
signs. Taylor et al.~\cite{TaylorFWSPB10} develop a convolutional
neighbourhood components analysis~\cite{GoldbergerRHS04} regression model, applied to head
and hand pose estimation.

{\bf Activity recognition from human pose analysis}: Higher-level analysis
tasks such as activity recognition can directly use human pose
estimation as input (e.g.~\cite{RamananF03}).  Models utilizing human
pose as a latent variable, estimated in the service of action
recognition, include Yao and Fei-Fei~\cite{yao2010modeling} and Yang et al.~\cite{YangWM10}
Indirect methods, looking at pixels to classify actions of people,
include Ikizler et al.~\cite{ikizler2009learning}, who learn
pose-based action from images obtained from internet searches.

{\bf Learning similarities}: Distance function learning is a
well-studied problem.  Early work by Xing et al.~\cite{XingNJR02} used
a set of similar pairs and a set of dissimilar pairs to formulate a
learning objective.  Schultz and Joachims~\cite{SchultzJ03} work with
relative comparisons of distances.  The neighbourhood conponents
analysis~\cite{GoldbergerRHS04} model learns to minimize
nearest-neighbour classification error.  This was extended to a
mixture of sparse distance measures by Hong et al.~\cite{HongLJT11}.
Weinberger and Saul~\cite{WeinbergerS09} similarily learn distances
for nearest neighbour, but with a large margin criterion.  Norouzi et
al.~\cite{NorouziFS12} learn Hamming distance in a transformed space.

Frome et al.~\cite{FromeSM06,frome2007learning} did pioneering work on using triplets for
learning distance functions.  This was extended with deep learning for
learning to categorize images by Wang et al.~\cite{WangSLRWPCW14}.
More broadly, novel distance learning methods have been applied to
vision tasks ranging from face analysis to generic image retrieval or matching
\cite{GuillauminVS10, TuytelaarsFSD11, MensinkVPC13, JainKG08}.

\section{Learning Pose Embeddings}
Our goal is to learn an embedding that places images of humans in
similar poses nearby.  We pose the problem as a triplet learning
problem, similar to~\cite{WangSLRWPCW14}.  Given a pose similarity score
$S(p_i, p_j)$, where $p_i$ and $p_j$ are two human poses, we want to
learn an embedding function $f(p)$ such that 
\beq
D(f(p_i), f(p_i^+)) < D(f(p_i), f(p_i^-)) \\
\mbox{ s.t. } S(p_i, p_i^+) > S(p_i, p_i^-)
\eeq
$D(f(p_i), f(p_j))$ is a distance measure in the embedding space.
In our work, we use the squared Euclidean distance:
\beq
D(f(p_i), f(p_j)) = \norm f(p_i) - f(p_j) \norm^2
\eeq
We call $t_i = (p_i, p_i^+, p_i^-)$ a triplet.  The triplet is used to
rank the ordering of the three poses, where $p_i$ is a query pose and 
$p_i^+$ is a more similar pose than $p_i^-$.

Similar to~\cite{WangSLRWPCW14}, we use the deep neural network framework to
learn the embedding.  The loss is the hinge loss defined on the
triplet $t_i = (p_i, p_i^+, p_i^-)$:
\beq
l(pi, p_i^+, p_i^-) = \max(0, g + D(f(p_i), f(p_i^+)) - D(f(p_i),
  f(p_i^-)))  \label{eq:hingeloss}
\eeq
where $g$ is the gap parameter.

The network architecture is similar to the ``inception'' architecture
proposed by~\cite{Szegedy}.  The output of the network is $L2$
normalized to produce an embedding of dimension $128$.  Each image of
the triplet is processed by the network in parallel.  The three
embeddings are evaluted using the hinge loss~\ref{eq:hingeloss}.

The details of the network structure are summarized in Tab.~\ref{tab:network}.  Input images are resized to
128x128 pixels.  The first network layer consists of 7x7 convolution with rectified linear unit (ReLU) activation,
max pooling over 3x3 patches with 2x2 stride, and local response normalization.
This is followed by 1x1 bottleneck ReLU units, 3x3 convolution, and local response normalization.
Subsequent layers perform in parallel four operations: 1x1, 3x3, and 5x5 bottleneck-convolution sequences, 
and spatial pooling.  Pooling is either L2 aggregation keeping a fixed resolution, or max pooling over a 3x3 patch
with 2x2 stride to aggregrate responses to a coarser-level representation.
Bottleneck-convolution sequences utilize 2x2 strides when paired with max pooling to maintain equal spatial resolution.

This embedding approach is very efficient at test time.  Each image is represented as a 128-dimensional vector embedding.  This representation permits efficient search via data structures such as KD trees or hashing approaches.


\begin{table*}
\begin{tabular}{|c|c|c|}
\hline
Type & Structure & Nodes \\
\hline
Input & 128x128 pixels & $128*128$ \\
Conv & 7x7 filters, 1x1 stride, ReLU & $64*128*128$ \\
Max pooling & 3x3 patch, 2x2 stride & $64*64*64$ \\
Local response normalization & & $64*64*64$ \\
Bottleneck & 1x1 ReLU & $64*64*64$ \\
Conv & 3x3 filters, 1x1 stride, ReLU & $192*64*64$ \\
Local response normalization & & $192*64*64$ \\
Max pooling & 3x3 patch, 2x2 stride & $192*32*32$ \\
Mixed & 1x1, 3x3, 5x5 bottleneck-convolution ReLU, max pooling & $256*32*32$ \\
Mixed & 1x1, 3x3, 5x5 bottleneck-convolution ReLU, L2 pooling & $320*32*32$ \\
Mixed & 1x1, 3x3, 5x5 bottleneck-convolution ReLU, max pooling & $640*16*16$ \\
Mixed & 1x1, 3x3, 5x5 bottleneck-convolution ReLU, L2 pooling & $640*16*16$ \\
Mixed & 1x1, 3x3, 5x5 bottleneck-convolution ReLU, L2 pooling & $640*16*16$ \\
Mixed & 1x1, 3x3, 5x5 bottleneck-convolution ReLU, L2 pooling & $640*16*16$ \\
Mixed & 1x1, 3x3, 5x5 bottleneck-convolution ReLU, L2 pooling & $640*16*16$ \\
Mixed & 1x1, 3x3, 5x5 bottleneck-convolution ReLU, max pooling & $1024*8*8$ \\
Mixed & 1x1, 3x3, 5x5 bottleneck-convolution ReLU, L2 pooling & $1024*8*8$ \\
Mixed & 1x1, 3x3, 5x5 bottleneck-convolution ReLU, max pooling & $1024*8*8$ \\
Average pooling & 5x5 filter, 1x1 stride & $1024*8*8$ \\
Embedding and normalization & full connections & 128 \\
\hline
\end{tabular}
\caption{Network structure.}
\label{tab:network}
\end{table*}

We also implemented a scheme to harvest hard negatives within each
mini-batch in the loss layer.  For $p_i$ and $p_i^+$ in each triplet,
we search through the minibatch to find a negative which is hard.  All other
images in a minibatch are considered as potential hard negatives for a given triplet.
A hard negative is sampled from these images based on the embedding of the current
network.

\section{Experiments}
We demonstrate the efficacy of our pose embedding method for retrieval.
In order to train a pose embedding model we require triplets
of human pose images -- anchor images paired with positive (similar) and negative (dissimilar) images.
A number of methods could be used to acquire such triplets, including human raters and relevance feedback from image search, among others.
However, in order to allow controlled experiments we use images of humans with labeled body joints
from the MPII Human Pose Dataset~\cite{andriluka14cvpr}.

The MPII Human Pose Dataset contains the most diverse set of human pose-labeled images currently available as a benchmark.
We utilize only the images marked as training from the MPII Human Pose Dataset -- the labels for the test images are not
distributed with the dataset to prevent tuning for the benchmark's main purpose of human pose estimation evaluation.

We extracted 19919 human pose images from the MPII Human Pose Dataset, the training images which contained full, valid annotation
of all body joints.  A subset of 10000 human pose images was used to train our models and the remainder used for
evaluation.


\subsection{Training a Pose Embedding}

We extract human pose images using the annotations provided in the MPII Human Pose Dataset.  Each cropped human pose image
is resized to 128x128 pixels.  In order to train our pose embedding model, we need to provide triplets.  The 2D image
locations of 16 body joints are provided in the dataset.  We define a distance between a pair of human pose images by measuring the
average Euclidean distance between body joints after aligning the pair of poses via a translation to match the 2D torso (root)
locations\footnote{We examined other measurements such as the PCP used in pose estimation, and qualitatively found simple Euclidean
distance to be a superior measure for pose similarity.}.

Given this distance measure between poses, triplets were extracted by
considering each one of the 10000 training images in turn.  For a
given ``anchor" training image sets of nearby ``positive'' and
dissimilar ``negative'' images are chosen.  The positive set is
constructed by first thresholding Euclidean distance on joints; all
images with a mean joint distance less than 7 pixels are chosen as
positive images.  We augment this set with the 2 closest training
images in order to account for poses that have larger pose variation
-- simpler poses such as standing people otherwise tend to dominate
the set of positive images. In sum, the positive set consists of the 2 closest images to each
anchor, regardless of their joint distance, plus all images within 7
pixels mean joint distance.

For the negative images, a set of up to 5000 images is chosen.  Again,
Euclidean distance on poses is thresholded, this time at greater than
15 pixels.  The closest 5000 images with Euclidean distance on pose
greater than this threshold are used as the negative set.
From this set of positives and negatives, all possible pairs were
constructed and used to form triplets.  This resulted in a set of
$\approx 20M$ triplets for training the model.

\begin{figure}
\includegraphics[trim=0cm 5cm 10cm 0cm, clip=true, width=0.5\textwidth]{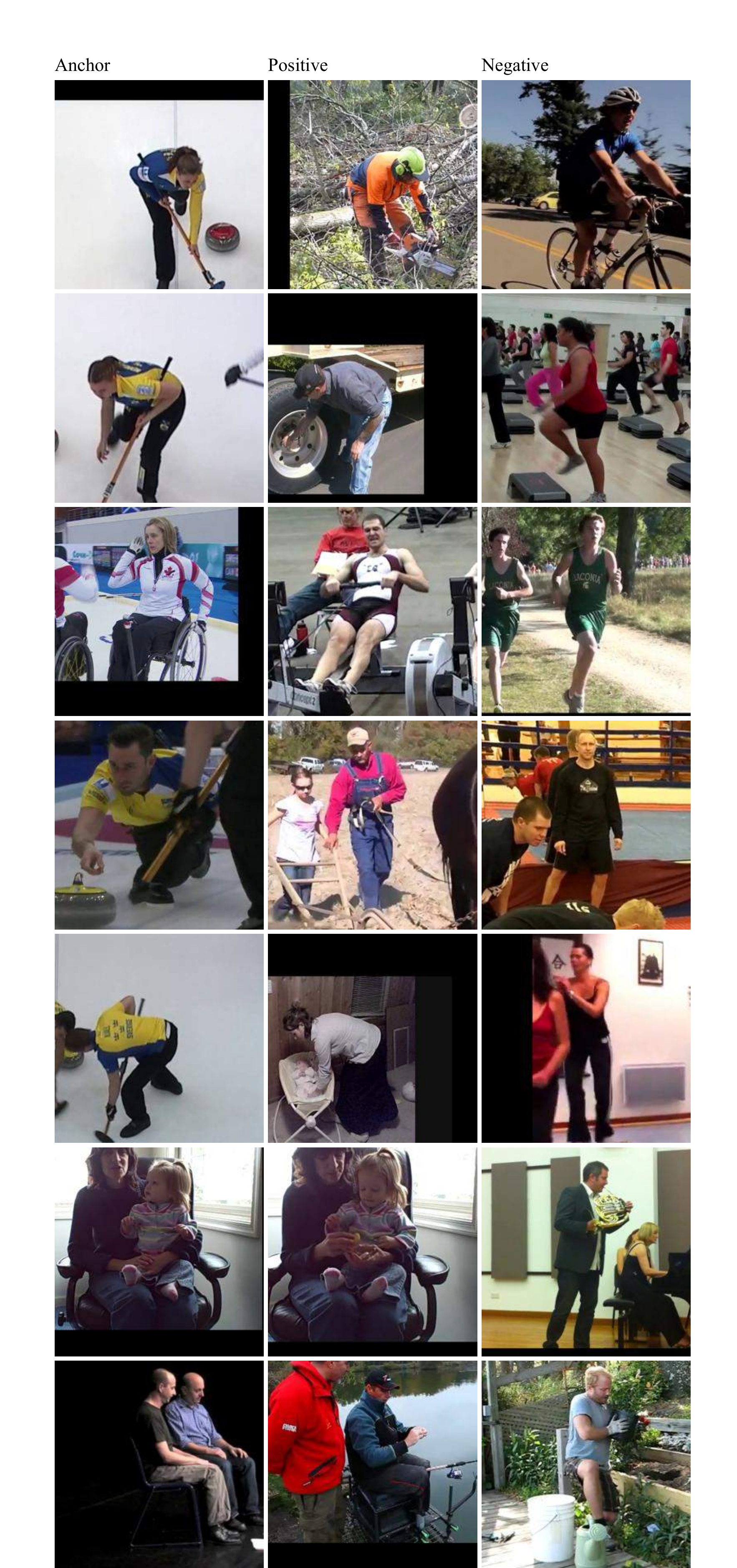}
\caption{Examples of triplets of poses used for training the pose
embedding.  First column is ``anchor'' image, second column is
``positive'' image of person in similar pose, third is ``negative''
image of person in a different pose.}
\label{fig:triplets}
\end{figure}

Examples of triplets are shown in Fig.~\ref{fig:triplets}.  Generally
these triplets capture qualitative pose similarity, though for heavily
foreshortened limbs 2d body joint locations do not necessarily lead to
good similarity measurements.  We believe that replacing this process
with a human rater or other means to derive similar-dissimilar labels
would result in higher quality training data, and at a potentially
lower cost than full body joint annotation.

We also used synthetic distortions of positive and anchor images to
increase the robustness of the learned model.  Images were rescaled to
a size uniformly sampled from the range $[0.9,1.1]$.  Translations,
again uniformly sampled $\pm 10\%$ of image size, were also randomly
chosen and applied to training images.

Network training was performed with random initialization of
parameters, batch size 600, AdaGrad with an initial of learning rate
0.05.

\subsection{Pose Retrieval Results}

We start by quantitatively evaluating the accuracy of our pose embedding by using it to perform pose retrieval on the MPII Human Pose Dataset.
The 9919 images not used for training are used for evaluation: 8000 are used as a set of known ``database" images, 1919 are used as query images.
For each of the query images, we use our pose embedding to find the most similar matches in the database images.
We compute $L_2$ distance between each query and each database image based on the embedding coordinates returned by our learned model.
We emphasize that the ground-truth human body joint locations are not used by our algorithm for any of these images, neither the
database images nor the query images.

In order to evaluate pose retrieval results we use three different performance measures.
\begin{itemize}
  \item {\bf Pose Difference}: We measure the pose difference between a query and each of the ranked images returned by the method.  Over a rank list
of length $K$, we again use Euclidean distance over body joints and find the best matching image in the rank list.

  \item {\bf Hit at $K$-absolute}: We define a threshold of 15 pixels in mean Euclidean joint distance as a ``correct" match between a
query image and a returned database image.  The Hit at $K$-absolute measure counts the fraction of query images that have at least
one correct match in a rank list of length $K$.

  \item {\bf Hit at $K$-relative}: We define the threshold to be relative to the best possible match in the database.  A ``correct" match between
a query image and a returned database image is one that is within $\tau+10$ pixels in mean Euclidean joint distance, where $\tau$ is the
closest database image to a given query image.
\end{itemize}

We present results using a variety of different models, in addition to
our pose embedding approach.
\begin{itemize}
  \item {\bf Oracle / random}: In order to gauge the difficulty of the
  retrieval problem, we also measure performance of an oracle and
  random selection.  The oracle method retrieves the closest matching
  image according to the ground truth joint positions.  The random
  method randomly chooses an image to match a query.

  \item {\bf ImageNet feature model}: This baseline uses generic image features obtained from a deep network trained for image
classification using the ImageNet dataset~\cite{ILSVRC15}.  We use an {\it Inception}~\cite{Szegedy} deep architecture, a model which obtains strong performance for image classification.
We take the penultimate layer of this network as a (1024 dimensional) feature vector to describe an image, and compare them using
Euclidean distance.

  \item {\bf Pose estimation model}: We train a model using the
  regression-based strategy for pose estimation in the Deep
  Pose~\cite{ToshevS14} approach as another baseline.  The network is
  a similar Inception architecture to that used in our pose embedding.
  The same training data as our approach, 10000 labeled MPII pose
  images, are used for training the pose estimation model.  Euclidean
  distance between 2d joint positions predicted by this model is used
  for retrieving images.  Note that this baseline requires detailed
  annotation of body joint locations for training, as opposed to our
  pose embedding method that only requires similar-dissimilar
  triplets.

  \item {\bf Combined pose embeddings with pose estimation}: We follow
  a straight-forward fusion strategy to combine our pose embeddings
  with the output of the Deep Pose pose estimation model.  The
  per-query distances returned by each method are normalized by the
  maximum distance to a database image, and the arithmetic mean of the
  two distances is used to fuse the distances.
\end{itemize}

\begin{figure*}
        \centering
        \begin{subfigure}[b]{0.3\textwidth}
                \includegraphics[trim=1cm 6cm 2cm 6cm, clip=true, width=\textwidth]{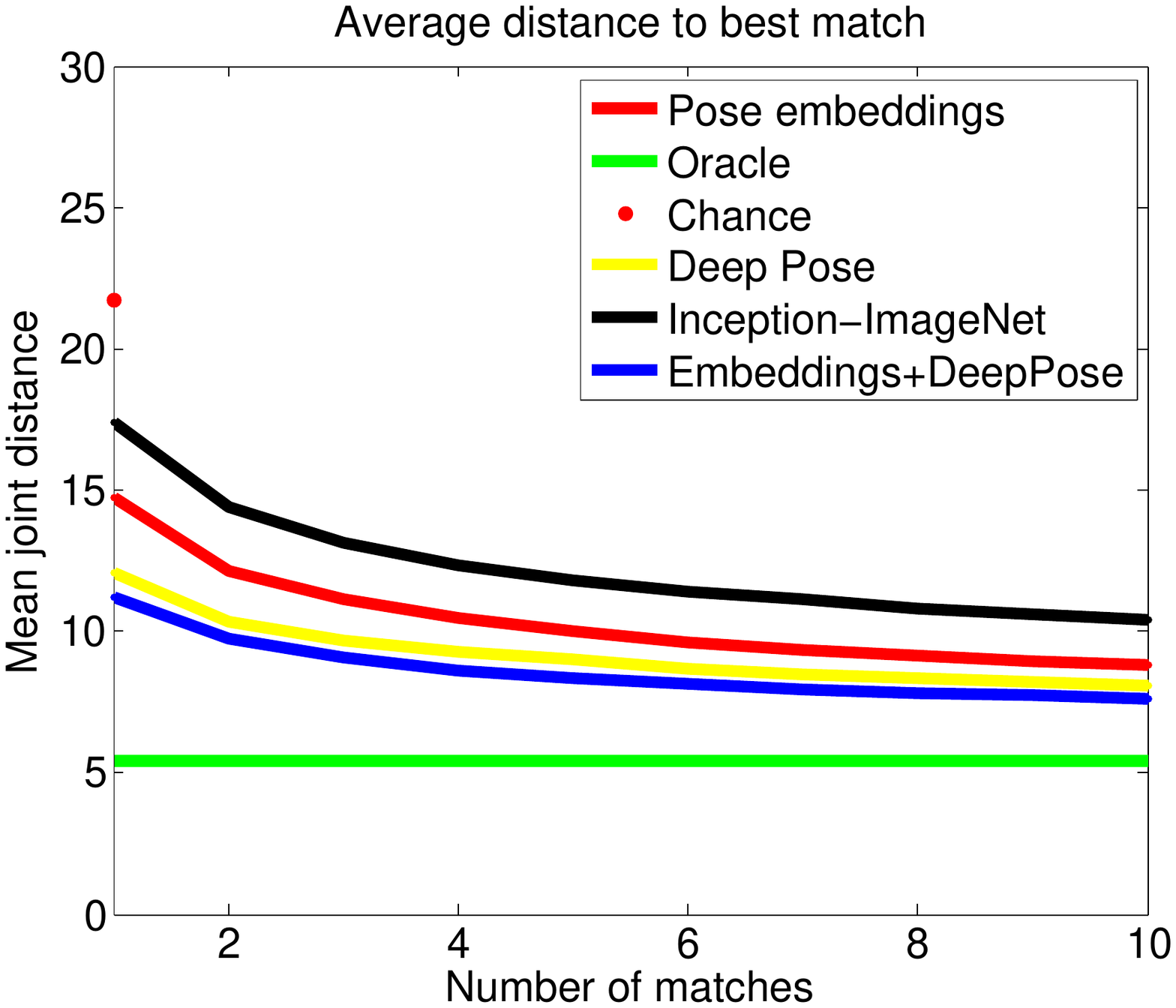}
                \caption{Average distance to nearest match.}
        \end{subfigure}
        \begin{subfigure}[b]{0.3\textwidth}
                \includegraphics[trim=1cm 6cm 2cm 6cm, clip=true, width=\textwidth]{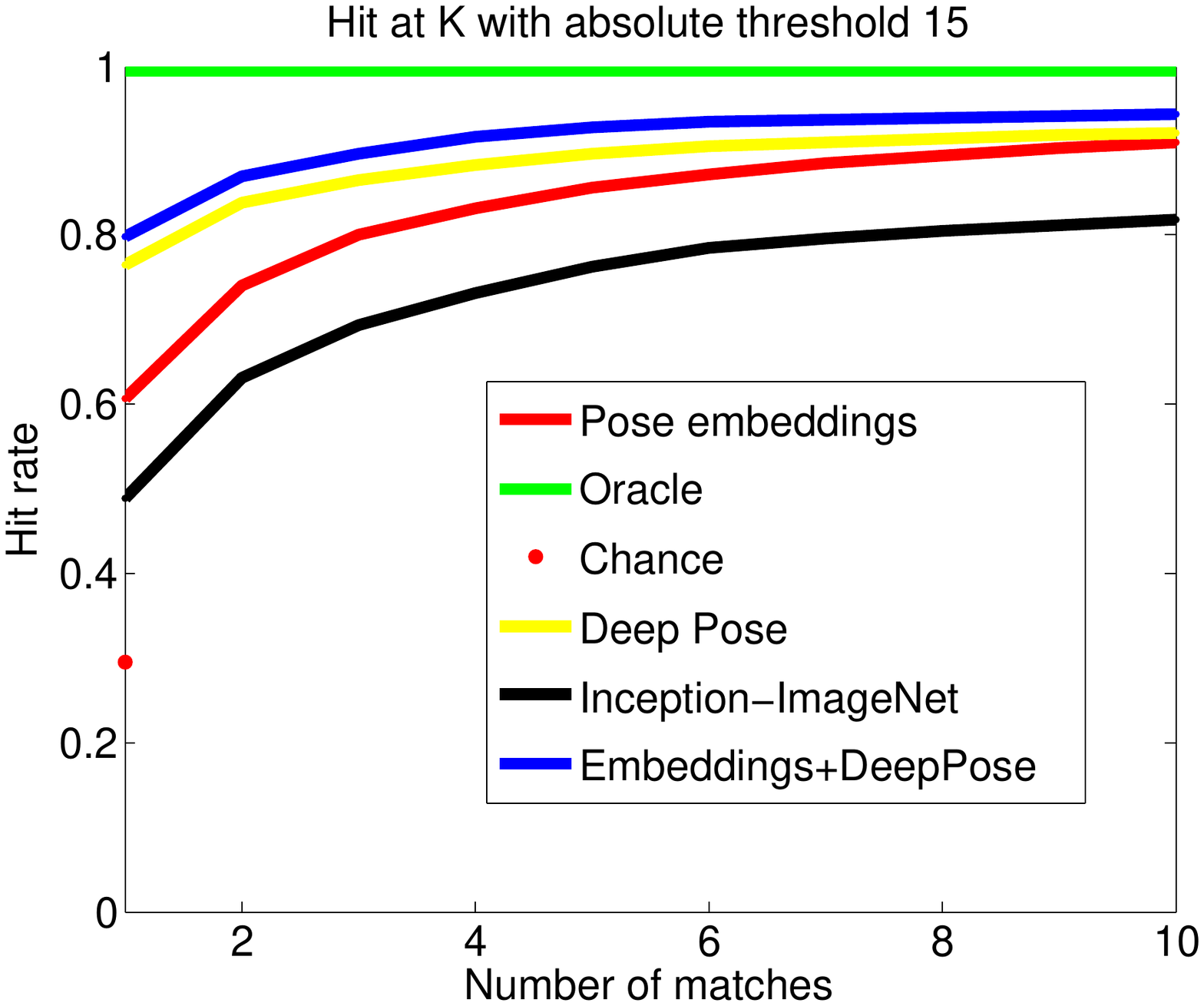}
                \caption{Hit at K-absolute.}
        \end{subfigure}
        \begin{subfigure}[b]{0.3\textwidth}
                \includegraphics[trim=1cm 6cm 2cm 6cm, clip=true, width=\textwidth]{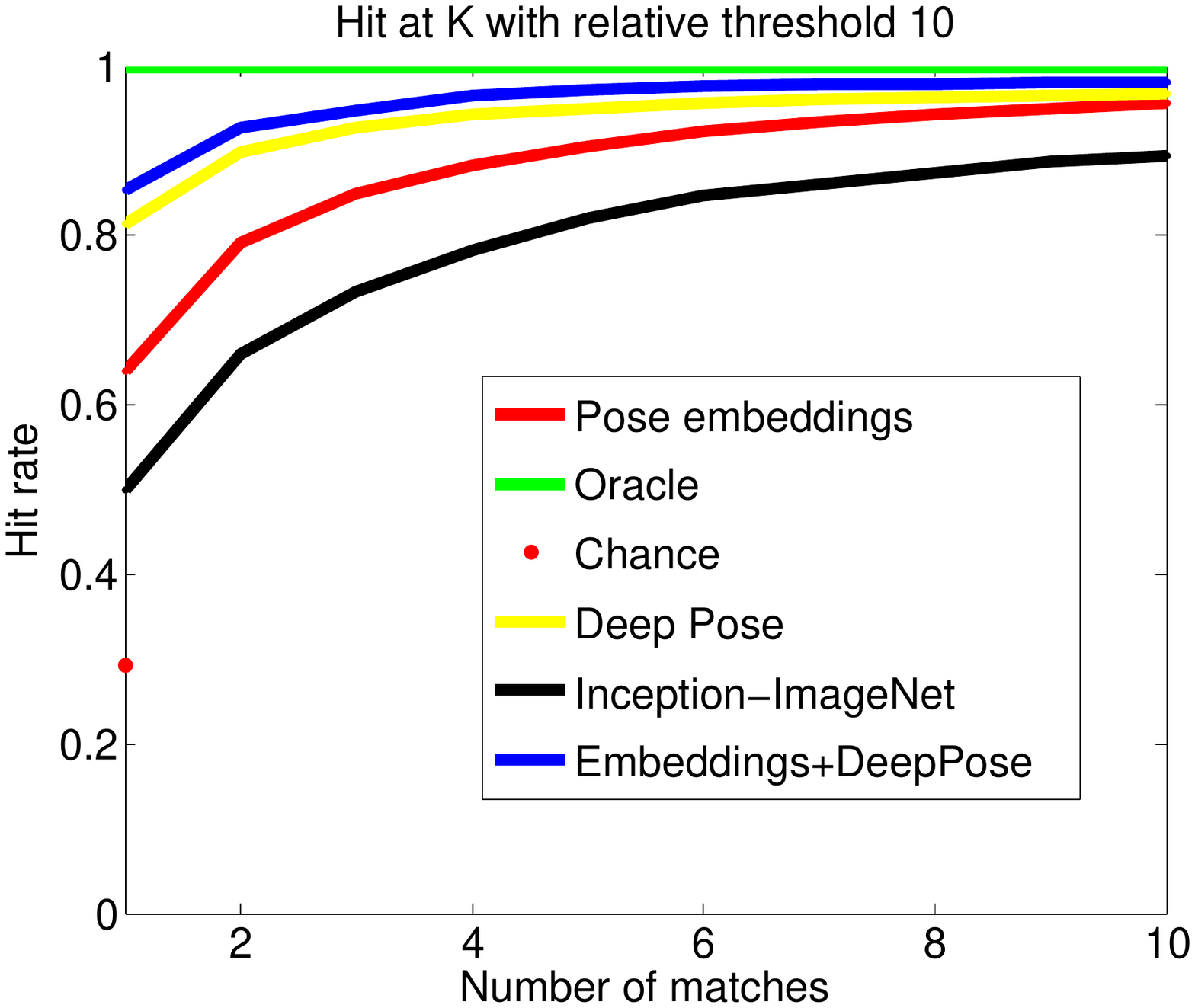}
                \caption{Hit at K-relative.}
        \end{subfigure}
\caption{Quantitative results on MPII Human Pose dataset.}
\label{fig:mpii_quant}
\end{figure*}

Quantitative results comparing our pose embedding method with these
baselines are presented in Fig.~\ref{fig:mpii_quant}.  We examine the
three performance measures across varying sizes of rank lists.  In
addition to the ImageNet-trained model and pose estimation results we
plot chance and oracle performance to provide context for the
difficulty of the task.  Our pose embedding model outperforms the
ImageNet models.  Qualitatively, as expected the ImageNet model
returns images with similar content (e.g.\ cyclists), rather than
focusing on human pose.

\begin{figure*}
\includegraphics[trim=0cm 4cm 0cm 0cm, clip=true, width=\textwidth]{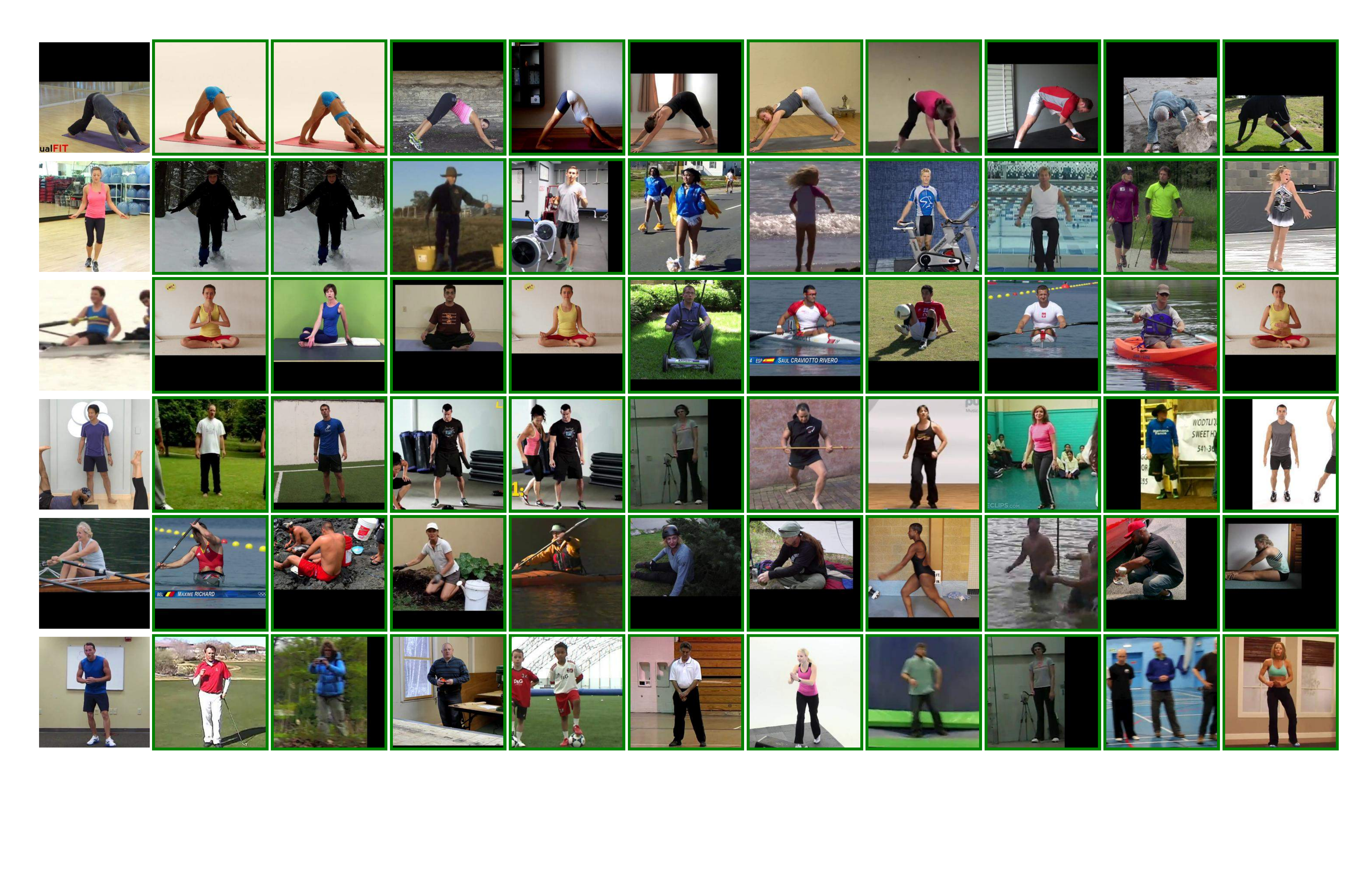}
\includegraphics[trim=0cm 4cm 0cm 1cm, clip=true, width=\textwidth]{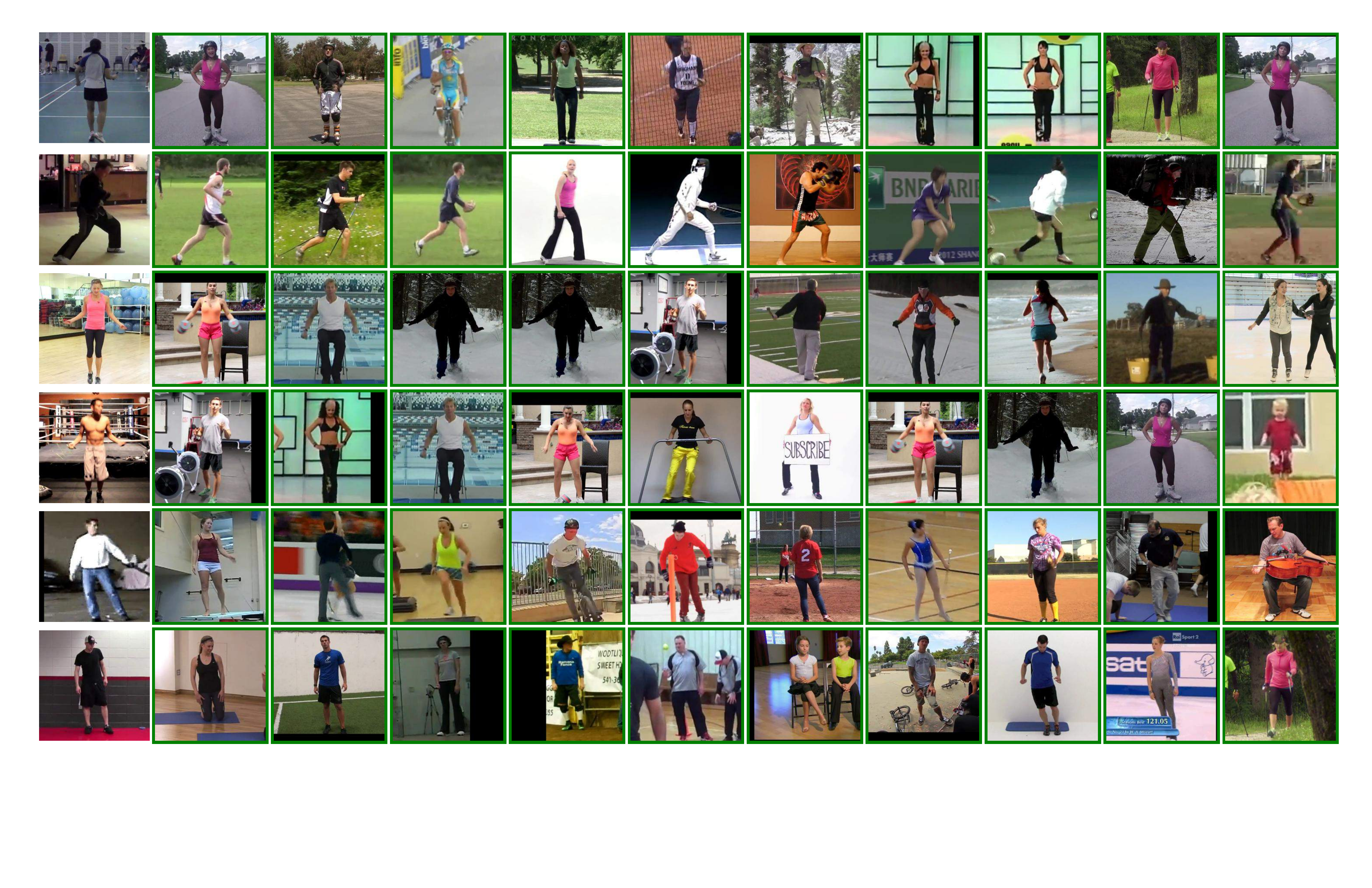}
\caption{Qualitative results of pose retrieval on MPII dataset.  The
        top 12 most confidently matched query images are shown.  First
column shows query image.  Subsequent columns show ranked list of
closest matching images based on learned pose embedding.}
\label{fig:mpii_qual}
\end{figure*}

The Deep Pose model outperforms the learned pose embeddings
quantitatively.  However, the proposed pose embedding method is
competitive, produces qualitatively very good retrieval results, and
can be used with less supervision.  Rank lists
from our pose embedding method are shown in Fig.~\ref{fig:mpii_qual}.
These queries are the most confidently matched in the test set: sorted
by distance to the nearest match.  Further, the information provided
by the pose embeddings is complementary to that of the Deep Pose
model; fusing these two leads to an improvement in quantitative performance.

\begin{figure*}
        \centering
        \begin{subfigure}[b]{0.48\textwidth}
                \includegraphics[trim=0cm 2cm 10cm 0cm, clip=true, width=\textwidth]{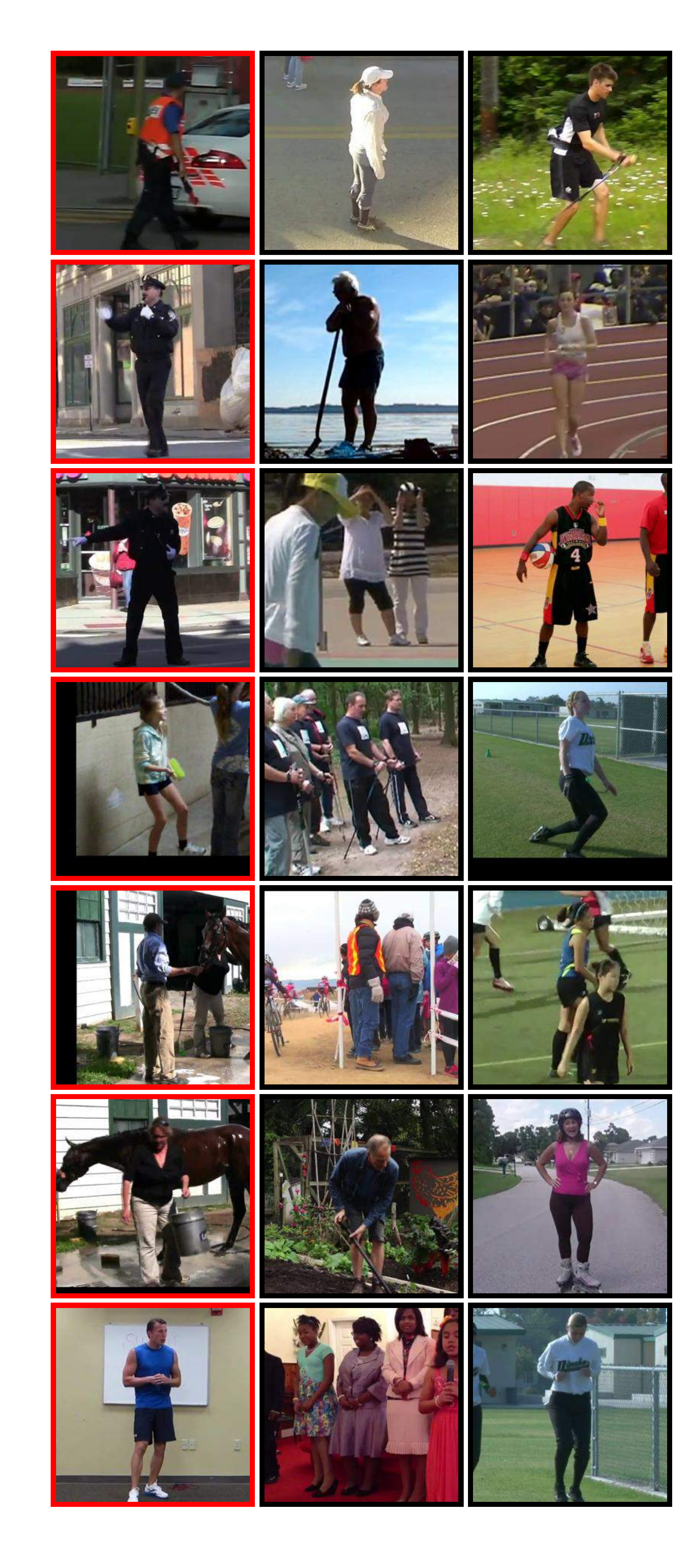}
        \end{subfigure}
        \begin{subfigure}[b]{0.48\textwidth}
                \includegraphics[trim=0cm 2cm 10cm 0cm, clip=true, width=\textwidth]{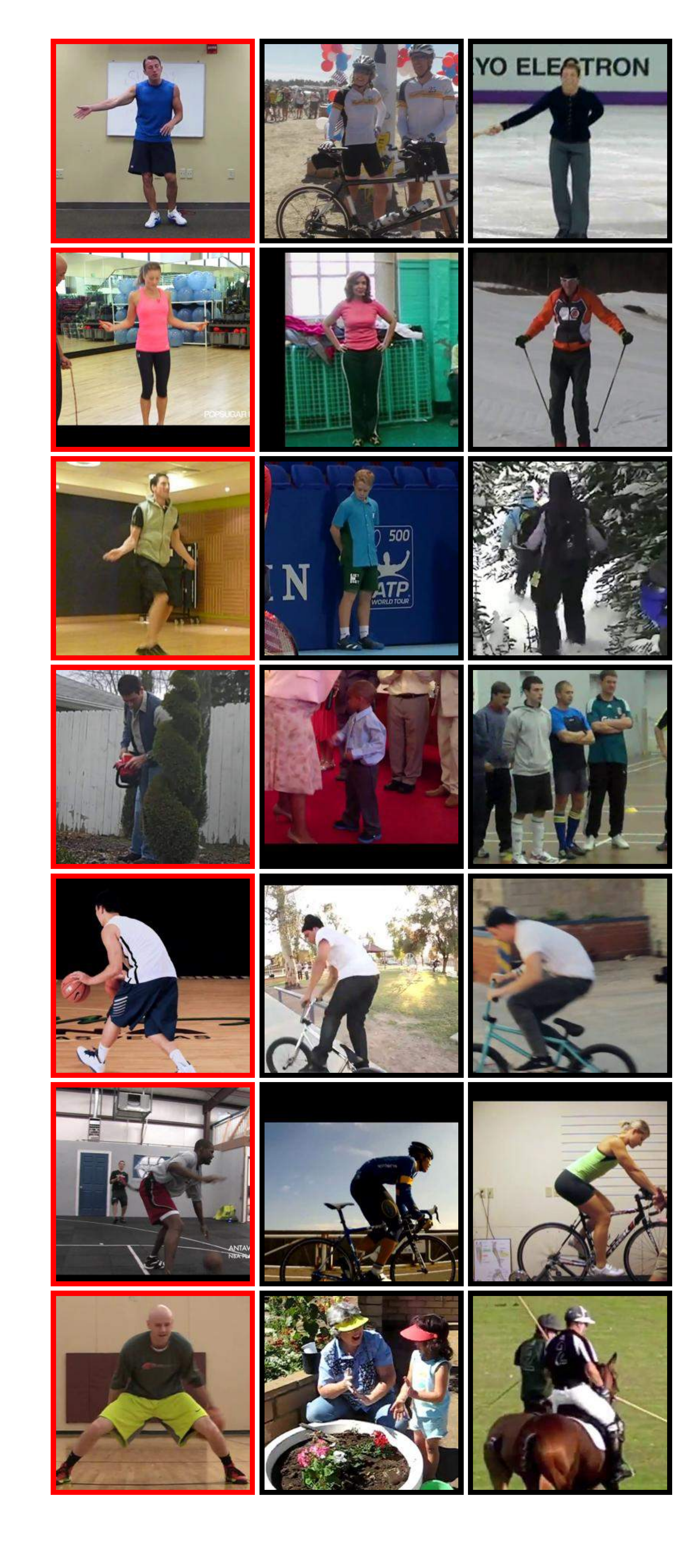}
        \end{subfigure}
\caption{Qualitative comparison of pose emebddings with similarity
        based on Deep Pose.  First column (red border) shows query
        image.  Second column is most similar image using Deep Pose.
        Third column is most similar image using learned pose embeddings.}
\label{fig:side-by-side}
\end{figure*}

Fig.~\ref{fig:side-by-side} shows qualitative comparisons of the
learned pose embeddings with similarity based on pose estimation using
Deep Pose~\cite{ToshevS14}.  Generally, both methods produce
qualitatively reasonable matches.  Deep pose performs very well, but
has occasional difficulty with unusual poses / occlusion.  Further, it requires
full labeled joint positions as training data, whereas our method can
be used with only similar-dissimilar labels.  A common error source
for the learned
pose embeddings is front-back flips (matching a person facing away
from the camera to one facing the camera).  Adding
additional training data or directly incorporating face
detection-based features could remedy these mistakes.

\section{Conclusion}
In this paper we presented a method for learning to match images based
on the pose of the person they contain.  The learning framework
utilizes triplets of images and learns a deep network that separates
similar images from dissimilar.  Since the learned
matching is based upon an embedding, it permits fast querying for
similar images.  We demonstrated the effectiveness of this framework
in pose matching. Future work includes applying this
framework with more general triplet similarity, for instance based on
image search relevance feedback or human ratings of triplet similarity.

{\small
\bibliographystyle{ieee}
\bibliography{egbib}
}

\end{document}